\documentclass[conference]{IEEEtran}
\IEEEoverridecommandlockouts
\usepackage{CJKutf8}
\usepackage{cite}
\usepackage{url}
\usepackage{amsmath,amssymb,amsfonts}
\usepackage{graphicx}
\usepackage{textcomp}
\usepackage{multicol}
\usepackage{multirow}
\usepackage{xcolor}
\usepackage{makecell}
\usepackage{changepage}
\usepackage{threeparttable}
\usepackage{hhline}
\usepackage{subfigure}
\usepackage{float}
\usepackage[ruled,vlined,linesnumbered]{algorithm2e}
\def\BibTeX{{\rm B\kern-.05em{\sc i\kern-.025em b}\kern-.08em
		T\kern-.1667em\lower.7ex\hbox{E}\kern-.125emX}}
\makeatletter
\newcommand{\linebreakand}{%
\end{@IEEEauthorhalign}
\hfill\mbox{}\par
\mbox{}\hfill\begin{@IEEEauthorhalign}
}
\makeatother

\begin{document}
\begin{CJK}{UTF8}{gbsn}

\title{RDEx-CSOP: Feasibility-Aware Reconstructed Differential Evolution with Adaptive $\varepsilon$-Constraint Ranking}

\author{
	\IEEEauthorblockN{
		Sichen Tao\textsuperscript{1,2},
		Yifei Yang\textsuperscript{3},
		Ruihan Zhao\textsuperscript{4,5},
		Kaiyu Wang\textsuperscript{6,7},
		Sicheng Liu\textsuperscript{8},
		Shangce Gao\textsuperscript{1}
	}
	\IEEEauthorblockA{\textsuperscript{1}Department of Engineering, University of Toyama, Toyama-shi 930-8555, Japan}
	\IEEEauthorblockA{\textsuperscript{2}Cyberscience Center, Tohoku University, Sendai-shi 980-8578, Japan}
	\IEEEauthorblockA{\textsuperscript{3}Faculty of Science and Technology, Hirosaki University, Hirosaki-shi 036-8560, Japan}
	\IEEEauthorblockA{\textsuperscript{4}Sino-German College of Applied Sciences, Tongji University, Shanghai 200092, China}
	\IEEEauthorblockA{\textsuperscript{5}School of Mechanical Engineering, Tongji University, Shanghai 200092, China}
	\IEEEauthorblockA{\textsuperscript{6}Chongqing Institute of Microelectronics Industry Technology,\\University of Electronic Science and Technology of China, Chongqing 401332, China}
	\IEEEauthorblockA{\textsuperscript{7}Artificial Intelligence and Big Data College,\\Chongqing Polytechnic University of Electronic Technology, Chongqing 401331, China}
	\IEEEauthorblockA{\textsuperscript{8}Department of Information Engineering, Yantai Vocational College, Yantai 264670, China}
	\IEEEauthorblockA{
		\{sichen.tao@tohoku.ac.jp, taosc73@hotmail.com\}; yyf7236@hirosaki-u.ac.jp;\\
		\{ruihan\_zhao@tongji.edu.cn, ruihan.z@outlook.com\};\\
		\{wangky@uestc.edu.cn, greskowky1996@163.com\};\\
		\{20250024@ytvc.edu.cn, lsctoyama2020@gmail.com\}; gaosc@eng.u-toyama.ac.jp
	}
}

\maketitle

\begin{abstract}
Constrained single-objective numerical optimisation requires both feasibility maintenance and strong objective-value convergence under limited evaluation budgets.
This report documents RDEx-CSOP, a constrained differential evolution variant used in the IEEE CEC 2025 numerical optimisation competition (C06 special session).
RDEx-CSOP combines success-history parameter adaptation with an exploitation-biased hybrid search and an $\varepsilon$-constraint handling mechanism with a time-varying threshold.
We evaluate RDEx-CSOP on the official CEC 2025 CSOP benchmark using the U-score framework (Speed, Accuracy, and Constraint categories).
The results show that RDEx-CSOP achieves the highest total score and the best average rank among all released comparison algorithms, mainly through strong speed and competitive constraint-handling performance across the 28 benchmark functions.
\end{abstract}


\section{Introduction}
Differential evolution (DE) and its adaptive descendants are among the most competitive optimisers for continuous real-parameter search because they combine strong directional variation with a lightweight algorithmic structure~\cite{storn1997differential,zhang2009jade,tanabe2013success,tanabe2014improving}. Reviewing studies on parameter control further show that the SHADE/L-SHADE line is one of the most reliable fixed-budget design families in modern DE~\cite{tanabe2020reviewing}. In unconstrained single-objective optimisation, this line has progressively improved search quality by strengthening parameter adaptation, selective pressure, and local perturbation, leading to variants such as LSHADE-RSP, iLSHADE-RSP, EB-LSHADE, and LSRTDE~\cite{stanovov2018lshade,choi2021improved,mohamed2019novel,stanovov2024successrate,stanovov2024lsrtde}. These studies collectively show that both operator design and parameter learning are decisive when the evaluation budget is tight.

Constrained single-objective optimisation adds a second layer of difficulty because the algorithm must not only improve objective values, but also reach and preserve feasibility efficiently. The constrained-optimisation literature has repeatedly shown that feasibility rules and $\varepsilon$-level control are not superficial add-ons: they redefine how candidate solutions are ranked, which solutions survive, and therefore what learning signal is available to an adaptive optimiser~\cite{deb2000efficient,takahama2006epsilon,mezura2011constraint}. As a result, a strong bound-constrained DE cannot simply be transplanted unchanged into constrained search; mutation bias, replacement, and parameter memories must all remain compatible with feasibility-aware ranking.

This issue becomes sharper in the CEC 2025 CSOP track, where algorithms are evaluated under the U-score framework rather than by final objective values alone~\cite{price2023trial}. In this regime, a method that is accurate but slow to attain feasibility, or fast but unstable after becoming feasible, can both be penalised. What is needed is therefore not only a strong DE operator set, but also a constraint-handling mechanism that shapes the search trajectory early enough to improve both feasible target-attainment speed and final solution quality.

RDEx-CSOP is designed from this perspective as a feasibility-aware reconstruction of the recent competition DE line. It integrates adaptive $\varepsilon$-ranking directly into selection and learning, while combining success-history memories, an exploitation-biased hybrid mutation branch, and success-rate-guided scaling-factor sampling. The main contribution is therefore a compact constrained DE framework in which operator adaptation is explicitly aligned with feasibility-aware search dynamics, with the goal of accelerating feasible progress without sacrificing final objective quality.
The source code for RDEx-CSOP is publicly available on Sichen Tao's GitHub page: \url{https://github.com/SichenTao}.

\section{Reconstructed Differential Evolution with $\varepsilon$-constraint Handling (RDEx-CSOP)}

\subsection{Problem Formulation}
RDEx-CSOP solves a constrained single-objective optimisation problem:
\begin{align}
\min_{x\in\mathbb{R}^D}\quad & f(x), \label{eq:csop_obj}\\
\text{s.t.}\quad
& g_i(x) \le 0,\quad i=1,\dots,m_g, \label{eq:csop_ineq}\\
& h_j(x) = 0,\quad j=1,\dots,m_h, \label{eq:csop_eq}\\
& \ell_k \le x_k \le u_k,\quad k=1,\dots,D. \label{eq:csop_bounds}
\end{align}

Following the released implementation, RDEx-CSOP uses the averaged violation measure
\begin{equation}\label{eq:csop_penalty}
\phi(x)=\frac{1}{m_g+m_h}
\begin{aligned}[t]
\Bigl(&\sum_{i=1}^{m_g}\max\{0,g_i(x)\}\\
&+\sum_{j=1}^{m_h}\max\{0,\lvert h_j(x)\rvert-\varepsilon_{\mathrm{eq}}\}\Bigr),
\end{aligned}
\end{equation}
where $\varepsilon_{\mathrm{eq}}=10^{-4}$ is the equality tolerance.

\subsection{Adaptive $\varepsilon$-ranking and Selection}
Let $\mathcal{F}^{(g)}=\{x_i^{(g)}\}_{i=1}^{N^{(g)}}$ denote the current elite front.
RDEx-CSOP sets the threshold from the empirical penalty distribution of $\mathcal{F}^{(g)}$, following the standard rationale of $\varepsilon$-level constrained search~\cite{takahama2006epsilon}:
\begin{equation}\label{eq:csop_eps_idx}
k^{(g)}=\max\!\left(1,\left\lfloor \eta N^{(g)}\left(1-\frac{\mathrm{NFE}}{\mathrm{MaxFE}}\right)^2\right\rfloor\right),\qquad \eta=0.8,
\end{equation}
\begin{equation}\label{eq:csop_epsilon}
\varepsilon^{(g)}=
\begin{cases}
\phi_{(k^{(g)})}^{(g)}, & \mathrm{NFE}\le 0.8\,\mathrm{MaxFE},\\
0, & \text{otherwise},
\end{cases}
\end{equation}
where $\phi_{(k)}^{(g)}$ is the $k$-th order statistic of $\{\phi(x_i^{(g)})\}_{i=1}^{N^{(g)}}$.
The corresponding ranking score is
\begin{equation}\label{eq:csop_rankscore}
s^{(g)}(x)=
\begin{cases}
f(x), & \phi(x)\le \varepsilon^{(g)},\\
f_{\max}^{(g)}+1+\phi(x), & \text{otherwise},
\end{cases}
\end{equation}
with $f_{\max}^{(g)}=\max_{x\in\mathcal{F}^{(g)}}f(x)$.
During one-to-one selection, let $\tilde{\phi}(x)=0$ if $\phi(x)\le\varepsilon^{(g)}$ and $\tilde{\phi}(x)=\phi(x)$ otherwise.
Then a trial $u_i^{(g)}$ replaces $x_i^{(g)}$ iff
\begin{equation}\label{eq:csop_select}
\begin{aligned}
\tilde{\phi}(u_i^{(g)})&<\tilde{\phi}(x_i^{(g)})\quad \text{or}\\
\tilde{\phi}(u_i^{(g)})&=\tilde{\phi}(x_i^{(g)})\land f(u_i^{(g)})\le f(x_i^{(g)}).
\end{aligned}
\end{equation}

\subsection{Standard Mutation and Crossover}
RDEx-CSOP maintains success-history memories $M_F$ and $M_{CR}$ of size $H$.
In the standard branch,
\begin{equation}\label{eq:csop_Fstd}
F_i^{(g)}\sim \mathcal{N}\!\left(\max\!\left(0,(SR^{(g)})^{1/3}\right),0.05^2\right)\cap[0,1],
\end{equation}
while $CR_i^{(g)}\sim \mathcal{N}(M_{CR,h},0.1^2)$ clipped to $[0,1]$.
The success-rate-guided centre in Eq.~(\ref{eq:csop_Fstd}) is motivated by the LSRTDE line~\cite{stanovov2024successrate,stanovov2024lsrtde}.
Using the $\varepsilon$-ranking in Eq.~(\ref{eq:csop_rankscore}), RDEx-CSOP samples $x_{pbest}^{(g)}$ from the best $p^{(g)}=\max(2,\lfloor 0.3N^{(g)}\rfloor)$ solutions and generates
\begin{equation}\label{eq:csop_std_mut}
v_i^{(g)} = x_i^{(g)} + F_i^{(g)}\!\left(x_{pbest}^{(g)}-x_i^{(g)}\right) + F_i^{(g)}\!\left(x_{r_1}^{(g)}-x_{r_2}^{(g)}\right),
\end{equation}
where $r_1$ is sampled from the front with an exponential rank bias and $r_2$ from the current population.
Binomial crossover is then applied:
\begin{equation}\label{eq:csop_xover}
u_{i,j}^{(g)}=
\begin{cases}
v_{i,j}^{(g)}, & \mathrm{rand}_j<CR_i^{(g)}\ \text{or}\ j=j_{\mathrm{rand}},\\
\hat{x}_{i,j}^{(g)}, & \text{otherwise},
\end{cases}
\end{equation}
where $\hat{x}_{i,j}^{(g)}$ equals either $x_{i,j}^{(g)}$ or a local perturbation $\mathrm{Cauchy}(x_{i,j}^{(g)},0.1)$ with probability $0.2$, following the local-refinement idea used in iLSHADE-RSP~\cite{choi2021improved}.
If a trial component violates the bounds, midpoint repair is used:
\begin{equation}\label{eq:csop_repair}
u_{i,j}^{(g)}\leftarrow
\begin{cases}
\frac{x_{i,j}^{(g)}+\ell_j}{2}, & u_{i,j}^{(g)}<\ell_j,\\
\frac{x_{i,j}^{(g)}+u_j}{2}, & u_{i,j}^{(g)}>u_j.
\end{cases}
\end{equation}

\subsection{Exploitation-biased Branch and Memory Update}
The exploitation-biased (EB) branch is activated more often in the later search stage through an adaptive rate $\rho_{\mathrm{EB}}^{(g)}$, following the ordered-donor intuition of EB-LSHADE~\cite{mohamed2019novel}.
After ranking three sampled donors by Eq.~(\ref{eq:csop_rankscore}) as $x_{\mathrm{best}}^{(g)}$, $x_{\mathrm{mid}}^{(g)}$, and $x_{\mathrm{worst}}^{(g)}$, the EB donor is
\begin{equation}\label{eq:csop_eb_mut}
v_i^{(g)} = x_i^{(g)} + F_i^{(g)}\!\left(x_{\mathrm{best}}^{(g)}-x_i^{(g)}\right) + F_i^{(g)}\!\left(x_{\mathrm{mid}}^{(g)}-x_{\mathrm{worst}}^{(g)}\right).
\end{equation}
In this branch, $F_i^{(g)}$ is sampled from a truncated Cauchy distribution centred at $M_{F,h}$ (or $0.4$ for the fallback slot), and $CR_i^{(g)}$ is sampled from $\mathcal{N}(M_{CR,h},0.1^2)$ (or from a fallback mean $0.9$) with early-stage lower bounds $CR_i^{(g)}\ge 0.7$ before $0.25\,\mathrm{MaxFE}$ and $CR_i^{(g)}\ge 0.6$ before $0.5\,\mathrm{MaxFE}$.

Let $\Delta_i$ denote the improvement contributed by a successful trial, and define $\Delta_{\mathrm{EB}}^{(g)}=\sum_{i\in\mathcal{S}_{\mathrm{EB}}^{(g)}}\Delta_i$ and $\Delta_{\mathrm{std}}^{(g)}=\sum_{i\in\mathcal{S}_{\mathrm{std}}^{(g)}}\Delta_i$.
The hybrid rate is updated by
\begin{equation}\label{eq:csop_rho}
\rho_{\mathrm{EB}}^{(g+1)}=
\begin{cases}
\dfrac{\Delta_{\mathrm{EB}}^{(g)}}{\Delta_{\mathrm{EB}}^{(g)}+\Delta_{\mathrm{std}}^{(g)}}, & \text{if }\Delta_{\mathrm{EB}}^{(g)},\Delta_{\mathrm{std}}^{(g)}>0,\\[1ex]
0.7, & \text{otherwise}.
\end{cases}
\end{equation}
If $A_i$ is the realised crossover ratio of trial $i$ and $w_i=\Delta_i/\sum_t\Delta_t$, then successful parameters update the memories through
\begin{equation}\label{eq:csop_memory}
\begin{aligned}
M_F &\leftarrow \frac{1}{2}\!\left(M_F+\frac{\sum_i w_i F_i^2}{\sum_i w_i F_i}\right),\\
M_{CR} &\leftarrow \frac{1}{2}\!\left(M_{CR}+\frac{\sum_i w_i A_i^2}{\sum_i w_i A_i}\right).
\end{aligned}
\end{equation}
The elite front is reduced linearly from $N_0$ to $N_{\min}=4$, again following the L-SHADE design principle~\cite{tanabe2014improving}:
\begin{equation}\label{eq:csop_lpsr}
N^{(g+1)}=\left\lfloor N_0+(N_{\min}-N_0)\frac{\mathrm{NFE}}{\mathrm{MaxFE}}\right\rfloor.
\end{equation}

\subsection{Pseudocode}
Algorithm~\ref{alg:rdex_csop} sketches the overall RDEx-CSOP procedure.
\begin{algorithm}[t]
\caption{RDEx-CSOP framework.}
\label{alg:rdex_csop}
\KwIn{$N_0$, memory size $H$, evaluation budget $\mathrm{MaxFE}$.}
\KwOut{Best feasible solution found.}
Initialise population/front, memories, and $\rho_{\mathrm{EB}}^{(0)}$\;
\While{$\mathrm{NFE}<\mathrm{MaxFE}$}{
  Compute $\varepsilon^{(g)}$ and $\varepsilon$-ranks by Eqs.~(\ref{eq:csop_epsilon})--(\ref{eq:csop_rankscore})\;
  \ForEach{$x_i$ in the current front}{
    Sample $(F_i,CR_i)$ from the memories\;
    Generate $v_i$ by Eq.~(\ref{eq:csop_std_mut}) or Eq.~(\ref{eq:csop_eb_mut})\;
    Apply crossover, perturbation, and repair by Eqs.~(\ref{eq:csop_xover})--(\ref{eq:csop_repair})\;
    Evaluate the trial and accept it by Eq.~(\ref{eq:csop_select})\;
  }
  Update $\rho_{\mathrm{EB}}$ and the memories by Eqs.~(\ref{eq:csop_rho}) and (\ref{eq:csop_memory})\;
  Reduce the front by Eq.~(\ref{eq:csop_lpsr})\;
}
\end{algorithm}

\section{Experimental Results}
\subsection{Benchmark Functions and Protocol}
The CEC 2025 CSOP track consists of 28 constrained problems with dimension $D=30$.
Each problem is evaluated with 25 independent runs and a maximum budget of $\mathrm{MaxFE}=20000\times D$ function evaluations.
The official U-score evaluation uses three categories: \textbf{Speed}, \textbf{Accuracy}, and \textbf{Constraint}, and aggregates results across problems under a fixed checkpoint schedule.

\subsection{Parameter Settings}
Unless otherwise stated, RDEx-CSOP uses the released reference configuration:
$N_0=600$, $H=5$, initial memories $M_F=0.3$ and $M_{CR}=1.0$, initial $\rho_{\mathrm{EB}}=0.7$, $\eta=0.8$ in Eq.~(\ref{eq:csop_eps_idx}), perturbation probability $0.2$, and the linear front reduction in Eq.~(\ref{eq:csop_lpsr}).

\subsection{Experimental Settings}
RDEx-CSOP is evaluated with the official U-score framework using the median target setting.
We compare RDEx-CSOP with all remaining algorithms available in the released competition package: RDEx, UDEIII, and CL-SRDE.
In the released evaluation files, the submitted winner is labelled as CORDEx; for naming consistency, we report it as RDEx-CSOP throughout this manuscript.

\subsection{Statistical Results}
\subsubsection{Overall U-score Results}
Table~\ref{tab:cec2025_csop_scores} reports the official median-target U-score results for all released comparison algorithms.
\begin{table}[t]
 \centering
 \caption{CEC 2025 CSOP evaluation (median target): overall scores over 28 problems and 25 runs for all released comparison algorithms.}
 \label{tab:cec2025_csop_scores}
 \scriptsize
 \renewcommand{\arraystretch}{0.95}
 \setlength{\tabcolsep}{2.5pt}
  \begin{tabular}{|c|l|r|r|r|r|r|}
   \hline
   Rank & Algorithm & Total Score & Avg Score/Prob. & Speed & Accuracy & Constraint \\ \hline
   1 & RDEx-CSOP & 53680.5 & 1917.16 & 44437.0 & 3440.0 & 5803.5 \\ \hline
   2 & RDEx & 49155.0 & 1755.54 & 39695.5 & 3730.5 & 5729.0 \\ \hline
   3 & UDEIII & 30545.0 & 1090.89 & 22368.5 & 3093.0 & 5083.5 \\ \hline
   4 & CL-SRDE & 28805.5 & 1028.77 & 19673.0 & 3245.5 & 5887.0 \\ \hline
  \end{tabular}
\end{table}

RDEx-CSOP achieves the highest total score ($53680.5$) and the best average rank ($1.36$) across all four released algorithms.
It leads the full field in the combined U-score and in the Speed category, while remaining competitive in the Accuracy and Constraint categories.
The earlier RDEx baseline ranks second with a total score of $49155.0$, confirming that the track-specific reconstruction further improves the official competition metric rather than only preserving the baseline level.

\subsubsection{Primary Statistical Tests}
Besides the official U-score, we report a single feasibility-aware final-quality indicator
\begin{equation}
Q_p(x)=
\begin{cases}
f(x), & \mathrm{CV}(x)\le 0,\\
B_p+\mathrm{CV}(x), & \mathrm{CV}(x)>0,
\end{cases}
\end{equation}
where $B_p$ is the largest finite final objective value on problem $p$ plus $1$.
This construction preserves the feasibility-first ordering and enables standard Wilcoxon and Friedman tests on one scalar per run.
Table~\ref{tab:csop_solid_summary} reports pairwise Wilcoxon W/T/L, Holm-corrected W/T/L, and median Vargha--Delaney $A_{12}$ values for $Q_p$ and TTT.
\begin{table}[t]
\centering
\caption{Primary pairwise summary over the 28 CEC2025 CSOP functions (25 runs). We report uncorrected per-function Wilcoxon W/T/L at $\alpha=0.05$, Holm-corrected W/T/L across functions, and the median Vargha--Delaney $A_{12}$ effect size for feasibility-aware final quality and time-to-target (larger is better for minimization).}
\label{tab:csop_solid_summary}
\scriptsize
\renewcommand{\arraystretch}{0.95}
\setlength{\tabcolsep}{2.5pt}
\begin{tabular}{|l|ccc|ccc|}
\hline
 \multirow{2}{*}{Competitor} & \multicolumn{3}{c|}{Final Q} & \multicolumn{3}{c|}{TTT} \\ \cline{2-7}
   & W/T/L & Holm & $A_{12}$ & W/T/L & Holm & $A_{12}$ \\ \hline
 RDEx & 1/26/1 & 1/27/0 & 0.52 & 4/24/0 & 1/27/0 & 0.58 \\ \hline
 UDEIII & 16/4/8 & 15/5/8 & 0.84 & 19/3/6 & 19/3/6 & 0.89 \\ \hline
 CL-SRDE & 6/18/4 & 6/19/3 & 0.50 & 24/4/0 & 23/5/0 & 0.92 \\ \hline
\end{tabular}
\end{table}

Against the earlier RDEx baseline, RDEx-CSOP is statistically comparable on final feasibility-aware quality ($1/26/1$) but faster on TTT ($4/24/0$).
Against UDEIII and CL-SRDE, RDEx-CSOP wins $16/4/8$ and $6/18/4$ functions on $Q_p$, and $19/3/6$ and $24/4/0$ functions on TTT, respectively.
Table~\ref{tab:csop_solid_friedman} further reports Friedman average ranks.
\begin{table}[t]
\centering
\caption{Primary Friedman tests on per-function medians over the 28 CEC2025 CSOP functions (25 runs). Final Q: $\chi^2=2.90$, $df=3$, $p=0.408$; TTT: $\chi^2=30.47$, $df=3$, $p=2.62E-06$. Lower average rank indicates better performance.}
\label{tab:csop_solid_friedman}
\scriptsize
\renewcommand{\arraystretch}{0.95}
\setlength{\tabcolsep}{2.5pt}
\begin{tabular}{|l|c|c|}
\hline
 Algorithm & Final Q & TTT \\ \hline
 RDEx-CSOP & \textbf{2.29} & \textbf{1.61} \\ \hline
 RDEx & 2.39 & 2.14 \\ \hline
 UDEIII & 2.84 & 2.89 \\ \hline
 CL-SRDE & 2.48 & 3.36 \\ \hline
\end{tabular}
\end{table}

The Friedman test is significant for TTT ($p=2.62\times 10^{-6}$) but not for $Q_p$ ($p=0.408$), indicating that the official U-score advantage is driven mainly by faster target attainment under competitive feasibility-aware final quality.

\subsubsection{Supplementary Diagnostics}
Because CSOP comparison is feasibility-first, split final-objective, split final-violation, and AUC analyses are treated only as complementary diagnostics.
Appendix~\ref{app:csop_tables} therefore separates supplementary solid statistical tables (full per-function $Q_p$ and TTT results) from a complementary-diagnostics section containing split objective, split violation, and AUC analyses.

\subsection{Time Complexity}
Let $D$ denote the problem dimension, $N$ the current front size, and $\mathrm{MaxFE}$ the evaluation budget.
Let $T_f$ be the average cost of evaluating both $f(x)$ and $\operatorname{cv}(x)$ once.
Per generation, RDEx-CSOP performs $O(ND)$ arithmetic operations for variation, repair, and bookkeeping, plus typically $O(N\log N)$ operations for ranking and front management.
Across a full run, the evaluation cost $O(\mathrm{MaxFE}\cdot T_f)$ dominates, while the algorithmic overhead is approximately $O(\mathrm{MaxFE}\cdot D)$ under the usual setting with $N=O(D)$.

\section{Conclusion}
This report presented RDEx-CSOP and its evaluation on the CEC 2025 CSOP benchmark suite.
The results demonstrate first-place official U-score performance against all released comparison algorithms and robust feasibility behaviour on most problems, supporting the effectiveness of combining success-history adaptation with an $\varepsilon$-constraint mechanism and exploitation-biased hybrid search.

\section*{Acknowledgment}
This research was partially supported by the Japan Society for the Promotion of Science (JSPS) KAKENHI under Grant JP22H03643, Japan Science and Technology Agency (JST) Support for Pioneering Research Initiated by the Next Generation (SPRING) under Grant JPMJSP2145, and JST through the Establishment of University Fellowships towards the Creation of Science Technology Innovation under Grant JPMJFS2115.

\bibliographystyle{IEEEtran}
\bibliography{references}

\clearpage
\onecolumn
\appendices
\section{Supplementary U-score Tables}
\label{app:csop_tables}
\begin{table}[H]
 \centering
 \caption{CEC 2025 CSOP evaluation (median target): average rankings over 28 problems (lower is better) for all released comparison algorithms.}
 \label{tab:cec2025_csop_ranks}
 \scriptsize
 \renewcommand{\arraystretch}{0.95}
 \setlength{\tabcolsep}{2.5pt}
  \begin{tabular}{|c|l|r|r|r|r|r|}
   \hline
   Rank & Algorithm & Total Rank & Avg Rank/Prob. & Avg Speed & Avg Accuracy & Avg Constraint \\ \hline
   1 & RDEx-CSOP & 38.0 & 1.36 & 1.59 & 2.75 & 2.52 \\ \hline
   2 & RDEx & 63.0 & 2.25 & 2.20 & 2.30 & 2.52 \\ \hline
   3 & UDEIII & 83.0 & 2.96 & 2.89 & 2.48 & 2.41 \\ \hline
   4 & CL-SRDE & 96.0 & 3.43 & 3.32 & 2.46 & 2.55 \\ \hline
  \end{tabular}
\end{table}

\section{Supplementary Solid Statistical Tables}
\begin{table}[H]
\centering
\caption{Feasibility-aware final-quality comparison on the 28 CEC2025 CSOP functions. For each run, the final-quality score equals the final objective/IGD value for feasible runs and $B_p+\mathrm{CV}$ for infeasible runs, where $B_p$ is the largest finite final objective/IGD value on problem $p$ plus $1$ (smaller is better).}
\label{tab:csop_quality_results}
\scriptsize
\renewcommand{\arraystretch}{0.9}
\setlength{\tabcolsep}{3pt}
\begin{tabular}{|c|cc|ccc|ccc|ccc|}
\hline
 \multirow{2}{*}{Problem} & \multicolumn{2}{c|}{RDEx-CSOP} & \multicolumn{3}{c|}{RDEx} & \multicolumn{3}{c|}{UDEIII} & \multicolumn{3}{c|}{CL-SRDE} \\ \cline{2-12}
   & Mean & SD & Mean & SD & W & Mean & SD & W & Mean & SD & W \\ \hline
 1 & \textbf{3.27E-30} & \textbf{4.75E-30} & \textbf{5.92E-30} & \textbf{9.27E-30} & = & \textbf{1.57E-28} & \textbf{9.70E-29} & + & \textbf{1.26E-31} & \textbf{6.18E-31} & = \\
 2 & \textbf{3.74E-30} & \textbf{5.62E-30} & \textbf{3.04E-30} & \textbf{4.86E-30} & = & \textbf{1.31E-28} & \textbf{7.83E-29} & + & \textbf{7.24E-31} & \textbf{2.42E-30} & = \\
 3 & 2.56E+02 & 7.36E+01 & 3.39E+02 & 1.63E+02 & = & \textbf{9.22E+01} & \textbf{5.54E+01} & - & 5.37E+02 & 1.07E+02 & + \\
 4 & 1.54E+01 & 1.42E+00 & 1.57E+01 & 1.45E+00 & = & \textbf{6.51E+00} & \textbf{6.78E+00} & - & 4.84E+01 & 8.86E+00 & + \\
 5 & \textbf{0.00E+00} & \textbf{0.00E+00} & \textbf{0.00E+00} & \textbf{0.00E+00} & = & \textbf{1.17E-28} & \textbf{4.17E-28} & = & \textbf{0.00E+00} & \textbf{0.00E+00} & = \\
 6 & \textbf{0.00E+00} & \textbf{0.00E+00} & 2.29E+00 & 1.00E+01 & = & \textbf{0.00E+00} & \textbf{0.00E+00} & = & 2.87E+00 & 1.41E+01 & = \\
 7 & \textbf{-1.39E+03} & \textbf{6.80E+02} & -8.70E+02 & 1.54E+02 & + & -6.73E+02 & 1.49E+02 & + & -8.28E+02 & 1.80E+02 & + \\
 8 & \textbf{-2.84E-04} & \textbf{0.00E+00} & \textbf{-2.84E-04} & \textbf{0.00E+00} & = & -2.84E-04 & 5.85E-12 & + & \textbf{-2.84E-04} & \textbf{0.00E+00} & = \\
 9 & \textbf{-2.67E-03} & \textbf{0.00E+00} & \textbf{-2.67E-03} & \textbf{0.00E+00} & = & -2.67E-03 & 4.34E-19 & + & \textbf{-2.67E-03} & \textbf{0.00E+00} & = \\
 10 & \textbf{-1.03E-04} & \textbf{0.00E+00} & \textbf{-1.03E-04} & \textbf{0.00E+00} & = & -1.03E-04 & 1.36E-20 & + & \textbf{-1.03E-04} & \textbf{0.00E+00} & = \\
 11 & 9.04E+00 & 3.65E+00 & 5.90E+00 & 5.49E+00 & - & 5.25E+01 & 1.23E+02 & + & \textbf{5.44E-02} & \textbf{3.05E+00} & - \\
 12 & 9.54E+00 & 1.14E+00 & 9.78E+00 & 0.00E+00 & = & \textbf{3.99E+00} & \textbf{2.18E-02} & - & 1.01E+01 & 4.05E+00 & = \\
 13 & \textbf{9.34E-29} & \textbf{2.14E-28} & \textbf{1.17E-28} & \textbf{2.34E-28} & = & 4.78E-01 & 1.30E+00 & + & \textbf{4.67E-29} & \textbf{1.58E-28} & = \\
 14 & 1.41E+00 & 0.00E+00 & 1.42E+00 & 2.36E-02 & = & \textbf{1.41E+00} & \textbf{0.00E+00} & - & 1.47E+00 & 4.05E-02 & + \\
 15 & \textbf{-3.93E+00} & \textbf{4.44E-16} & -3.93E+00 & 1.18E-05 & = & 2.36E+00 & 1.41E-06 & + & -3.42E+00 & 1.15E+00 & = \\
 16 & 1.99E+01 & 4.09E+00 & 2.10E+01 & 3.69E+00 & = & \textbf{0.00E+00} & \textbf{0.00E+00} & - & 2.32E+01 & 3.47E+00 & + \\
 17 & 3.35E+01 & 0.00E+00 & 3.35E+01 & 0.00E+00 & = & \textbf{3.19E+01} & \textbf{7.89E-01} & - & 3.35E+01 & 0.00E+00 & = \\
 18 & \textbf{3.65E+01} & \textbf{0.00E+00} & 3.65E+01 & 1.96E-05 & = & 8.00E+03 & 2.49E+03 & + & \textbf{3.65E+01} & \textbf{0.00E+00} & = \\
 19 & \textbf{4.27E+04} & \textbf{0.00E+00} & \textbf{4.27E+04} & \textbf{0.00E+00} & = & 4.28E+04 & 7.28E-12 & + & \textbf{4.27E+04} & \textbf{0.00E+00} & = \\
 20 & \textbf{1.26E+00} & \textbf{2.11E-01} & 1.33E+00 & 2.46E-01 & = & 1.85E+00 & 2.82E-01 & + & 2.43E+00 & 6.99E-01 & + \\
 21 & 2.16E+01 & 8.90E+00 & 2.39E+01 & 7.92E+00 & = & \textbf{9.28E+00} & \textbf{8.34E+00} & - & 2.47E+01 & 1.54E+01 & = \\
 22 & \textbf{3.75E-26} & \textbf{1.49E-26} & \textbf{5.61E-26} & \textbf{3.84E-26} & = & 2.56E+01 & 4.84E+01 & + & \textbf{2.85E-26} & \textbf{6.52E-27} & - \\
 23 & \textbf{1.41E+00} & \textbf{0.00E+00} & \textbf{1.41E+00} & \textbf{0.00E+00} & = & 1.45E+00 & 4.34E-02 & = & \textbf{1.41E+00} & \textbf{0.00E+00} & = \\
 24 & \textbf{-3.93E+00} & \textbf{4.44E-16} & \textbf{-3.93E+00} & \textbf{4.44E-16} & = & 2.36E+00 & 9.80E-08 & + & \textbf{-3.93E+00} & \textbf{4.44E-16} & = \\
 25 & 2.25E+01 & 3.00E+00 & 2.36E+01 & 3.69E+00 & = & \textbf{2.51E-01} & \textbf{1.23E+00} & - & 2.46E+01 & 3.10E+00 & = \\
 26 & 3.30E+01 & 0.00E+00 & 3.30E+01 & 0.00E+00 & = & \textbf{3.26E+01} & \textbf{8.09E-01} & = & 3.30E+01 & 0.00E+00 & = \\
 27 & 3.65E+01 & 8.06E-05 & 3.65E+01 & 8.00E-05 & = & 1.46E+04 & 5.11E+03 & + & \textbf{3.65E+01} & \textbf{2.71E-05} & - \\
 28 & 4.29E+04 & 1.14E+01 & 4.29E+04 & 1.10E+01 & = & 4.30E+04 & 1.47E+01 & + & \textbf{4.29E+04} & \textbf{7.28E-12} & - \\
\hline
 W/T/L & \multicolumn{2}{c|}{$-/-/-$} & \multicolumn{3}{c|}{1/26/1} & \multicolumn{3}{c|}{16/4/8} & \multicolumn{3}{c|}{6/18/4} \\
\hline
\end{tabular}
\end{table}

\begin{table}[H]
\centering
\caption{Time-to-target comparison on the 28 CEC2025 CSOP functions. TTT is the first checkpoint index where the run reaches the median target (smaller is better); runs that never reach the target are assigned 2001.}
\label{tab:csop_ttt_results}
\scriptsize
\renewcommand{\arraystretch}{0.9}
\setlength{\tabcolsep}{3pt}
\begin{tabular}{|c|cc|ccc|ccc|ccc|}
\hline
 \multirow{2}{*}{Problem} & \multicolumn{2}{c|}{RDEx-CSOP} & \multicolumn{3}{c|}{RDEx} & \multicolumn{3}{c|}{UDEIII} & \multicolumn{3}{c|}{CL-SRDE} \\ \cline{2-12}
   & Mean & SD & Mean & SD & W & Mean & SD & W & Mean & SD & W \\ \hline
 1 & 1332.6 & 546.2 & 1382.6 & 548.3 & = & 1979.6 & 105.0 & + & \textbf{1327.4} & \textbf{137.8} & = \\
 2 & \textbf{1251.2} & \textbf{514.8} & 1298.5 & 527.0 & = & 1960.7 & 138.2 & + & 1399.3 & 222.3 & + \\
 3 & 1575.3 & 541.4 & 1813.9 & 272.3 & = & \textbf{394.8} & \textbf{162.1} & - & 2001.0 & 0.0 & + \\
 4 & 1607.4 & 529.8 & 1569.9 & 496.5 & = & \textbf{837.6} & \textbf{50.4} & - & 2001.0 & 0.0 & + \\
 5 & \textbf{986.2} & \textbf{12.0} & 986.6 & 13.9 & = & 1476.4 & 466.1 & + & 1395.2 & 14.8 & + \\
 6 & \textbf{577.4} & \textbf{10.3} & 807.2 & 521.1 & = & 746.3 & 37.6 & + & 991.0 & 206.4 & + \\
 7 & \textbf{334.7} & \textbf{369.4} & 1130.0 & 739.4 & + & 1906.6 & 320.2 & + & 1606.0 & 480.3 & + \\
 8 & \textbf{765.0} & \textbf{13.3} & 769.6 & 11.7 & = & 2001.0 & 0.0 & + & 1221.0 & 7.6 & + \\
 9 & \textbf{50.5} & \textbf{12.9} & 56.2 & 12.3 & = & 1322.4 & 905.8 & + & 111.6 & 26.5 & + \\
 10 & \textbf{793.0} & \textbf{12.9} & 793.3 & 11.6 & = & 2001.0 & 0.0 & + & 1231.7 & 9.0 & + \\
 11 & 1.7 & 0.6 & \textbf{1.5} & \textbf{0.6} & = & 1.7 & 0.8 & = & 1.9 & 1.1 & = \\
 12 & 574.1 & 258.7 & 580.4 & 150.6 & = & \textbf{269.9} & \textbf{33.2} & - & 843.6 & 257.9 & + \\
 13 & \textbf{1172.4} & \textbf{361.7} & 1216.1 & 392.6 & = & 1672.0 & 383.1 & + & 1484.0 & 152.8 & + \\
 14 & \textbf{421.9} & \textbf{55.6} & 541.9 & 431.8 & = & 1036.1 & 2.7 & + & 1618.0 & 600.7 & + \\
 15 & \textbf{59.9} & \textbf{13.7} & 70.0 & 13.2 & + & 2001.0 & 0.0 & + & 148.7 & 25.8 & + \\
 16 & 1243.6 & 732.8 & 1468.6 & 667.3 & = & \textbf{131.8} & \textbf{71.2} & - & 1926.3 & 212.1 & + \\
 17 & 842.0 & 807.3 & 1232.2 & 869.8 & = & \textbf{212.0} & \textbf{197.6} & - & 1603.7 & 643.4 & + \\
 18 & \textbf{200.1} & \textbf{8.3} & 201.6 & 6.0 & = & 2001.0 & 0.0 & + & 307.7 & 7.1 & + \\
 19 & \textbf{280.2} & \textbf{25.9} & 290.6 & 40.3 & = & 311.4 & 49.3 & + & 560.7 & 254.5 & + \\
 20 & \textbf{1078.6} & \textbf{444.5} & 1084.2 & 472.0 & = & 1965.8 & 76.0 & + & 1922.5 & 215.4 & + \\
 21 & \textbf{396.5} & \textbf{173.5} & 498.9 & 341.1 & = & 464.7 & 455.1 & = & 800.5 & 556.8 & + \\
 22 & \textbf{1525.7} & \textbf{421.6} & 1568.4 & 415.9 & = & 2001.0 & 0.0 & + & 1560.4 & 130.4 & = \\
 23 & \textbf{433.6} & \textbf{28.0} & 452.5 & 30.1 & + & 1851.7 & 202.0 & + & 708.2 & 13.4 & + \\
 24 & \textbf{60.7} & \textbf{14.2} & 68.6 & 13.6 & = & 2001.0 & 0.0 & + & 141.8 & 21.9 & + \\
 25 & 1315.1 & 782.3 & 1459.0 & 727.9 & = & \textbf{515.5} & \textbf{424.5} & - & 1857.0 & 304.0 & = \\
 26 & 1102.0 & 803.1 & 1206.8 & 806.3 & = & \textbf{826.8} & \textbf{873.9} & = & 1772.2 & 460.9 & + \\
 27 & 218.8 & 34.6 & \textbf{216.3} & \textbf{13.5} & = & 2001.0 & 0.0 & + & 324.0 & 7.7 & + \\
 28 & \textbf{263.0} & \textbf{12.8} & 272.4 & 11.6 & + & 2001.0 & 0.0 & + & 459.6 & 21.2 & + \\
\hline
 W/T/L & \multicolumn{2}{c|}{$-/-/-$} & \multicolumn{3}{c|}{4/24/0} & \multicolumn{3}{c|}{19/3/6} & \multicolumn{3}{c|}{24/4/0} \\
\hline
\end{tabular}
\end{table}

\section{Complementary Diagnostics}
The tables in this section are diagnostic supplements to the official U-score results. For constrained tracks, split objective/IGD and split final-CV comparisons are not treated as independent primary criteria.
\begin{table}[H]
\centering
\caption{Complementary pairwise summary over the 28 CEC2025 CSOP functions (25 runs). For each metric (Final Obj., Final CV, and AUC), we report uncorrected per-function Wilcoxon W/T/L at $\alpha=0.05$, Holm-corrected W/T/L across functions, and the median Vargha--Delaney $A_{12}$ effect size (larger is better for minimization).}
\label{tab:csop_summary}
\scriptsize
\renewcommand{\arraystretch}{0.95}
\setlength{\tabcolsep}{2.5pt}
\begin{tabular}{|l|ccc|ccc|ccc|}
\hline
 \multirow{2}{*}{Competitor} & \multicolumn{3}{c|}{Final Obj.} & \multicolumn{3}{c|}{Final CV} & \multicolumn{3}{c|}{AUC} \\ \cline{2-10}
   & W/T/L & Holm & $A_{12}$ & W/T/L & Holm & $A_{12}$ & W/T/L & Holm & $A_{12}$ \\ \hline
 RDEx & 1/27/0 & 1/27/0 & 0.54 & 1/26/1 & 0/28/0 & 0.50 & 1/27/0 & 1/27/0 & 0.55 \\ \hline
 UDEIII & 14/7/7 & 14/7/7 & 0.73 & 22/5/1 & 21/6/1 & 1.00 & 15/4/9 & 14/6/8 & 0.73 \\ \hline
 CL-SRDE & 10/16/2 & 10/17/1 & 0.51 & 2/22/4 & 2/22/4 & 0.50 & 28/0/0 & 28/0/0 & 1.00 \\ \hline
\end{tabular}
\end{table}

\begin{table}[H]
\centering
\caption{Complementary Friedman tests on per-function medians over the 28 CEC2025 CSOP functions (25 runs). Final Obj.: $\chi^2=4.25$, $df=3$, $p=0.234$; Final CV: $\chi^2=29.00$, $df=3$, $p=4.82E-06$; AUC: $\chi^2=39.13$, $df=3$, $p=7.74E-08$. Lower average rank indicates better performance for each metric.}
\label{tab:csop_friedman_summary}
\scriptsize
\renewcommand{\arraystretch}{0.95}
\setlength{\tabcolsep}{2.5pt}
\begin{tabular}{|l|c|c|c|}
\hline
 Algorithm & Final Obj. & Final CV & AUC \\ \hline
 RDEx-CSOP & \textbf{2.11} & 2.21 & \textbf{1.75} \\ \hline
 RDEx & 2.46 & 2.20 & 1.96 \\ \hline
 UDEIII & 2.66 & 3.62 & 2.57 \\ \hline
 CL-SRDE & 2.77 & \textbf{1.96} & 3.71 \\ \hline
\end{tabular}
\end{table}

\begin{table}[H]
\centering
\caption{Final objective comparison on the 28 CEC2025 CSOP functions. For each algorithm, the mean and SD over 25 runs are reported; $W$ gives the Wilcoxon outcome of RDEx-CSOP against the competitor.}
\label{tab:csop_obj_results}
\scriptsize
\renewcommand{\arraystretch}{0.9}
\setlength{\tabcolsep}{3pt}
\begin{tabular}{|c|cc|ccc|ccc|ccc|}
\hline
 \multirow{2}{*}{Problem} & \multicolumn{2}{c|}{RDEx-CSOP} & \multicolumn{3}{c|}{RDEx} & \multicolumn{3}{c|}{UDEIII} & \multicolumn{3}{c|}{CL-SRDE} \\ \cline{2-12}
   & Mean & SD & Mean & SD & W & Mean & SD & W & Mean & SD & W \\ \hline
 1 & \textbf{3.27E-30} & \textbf{4.75E-30} & \textbf{5.92E-30} & \textbf{9.27E-30} & = & \textbf{1.57E-28} & \textbf{9.70E-29} & + & \textbf{1.26E-31} & \textbf{6.18E-31} & = \\
 2 & \textbf{3.74E-30} & \textbf{5.62E-30} & \textbf{3.04E-30} & \textbf{4.86E-30} & = & \textbf{1.31E-28} & \textbf{7.83E-29} & + & \textbf{7.24E-31} & \textbf{2.42E-30} & = \\
 3 & 2.56E+02 & 7.36E+01 & 3.39E+02 & 1.63E+02 & = & \textbf{9.22E+01} & \textbf{5.54E+01} & - & 5.37E+02 & 1.07E+02 & + \\
 4 & 1.54E+01 & 1.42E+00 & 1.57E+01 & 1.45E+00 & = & \textbf{6.51E+00} & \textbf{6.78E+00} & - & 4.84E+01 & 8.86E+00 & + \\
 5 & \textbf{0.00E+00} & \textbf{0.00E+00} & \textbf{0.00E+00} & \textbf{0.00E+00} & = & \textbf{1.17E-28} & \textbf{4.17E-28} & = & \textbf{0.00E+00} & \textbf{0.00E+00} & = \\
 6 & \textbf{0.00E+00} & \textbf{0.00E+00} & 2.29E+00 & 1.00E+01 & = & \textbf{0.00E+00} & \textbf{0.00E+00} & = & 2.87E+00 & 1.41E+01 & = \\
 7 & \textbf{-1.39E+03} & \textbf{6.80E+02} & -8.70E+02 & 1.54E+02 & + & -6.73E+02 & 1.49E+02 & + & -8.28E+02 & 1.80E+02 & + \\
 8 & \textbf{-2.84E-04} & \textbf{0.00E+00} & \textbf{-2.84E-04} & \textbf{0.00E+00} & = & -2.84E-04 & 5.85E-12 & + & \textbf{-2.84E-04} & \textbf{0.00E+00} & = \\
 9 & \textbf{-2.67E-03} & \textbf{0.00E+00} & \textbf{-2.67E-03} & \textbf{0.00E+00} & = & -2.67E-03 & 4.34E-19 & + & \textbf{-2.67E-03} & \textbf{0.00E+00} & = \\
 10 & \textbf{-1.03E-04} & \textbf{0.00E+00} & \textbf{-1.03E-04} & \textbf{0.00E+00} & = & -1.03E-04 & 1.36E-20 & + & \textbf{-1.03E-04} & \textbf{0.00E+00} & = \\
 11 & -7.87E+00 & 5.44E+00 & -5.09E+00 & 4.57E+00 & = & \textbf{-1.98E+02} & \textbf{4.90E+02} & = & -1.59E+00 & 2.54E+00 & + \\
 12 & 9.54E+00 & 1.14E+00 & 9.78E+00 & 0.00E+00 & = & \textbf{3.99E+00} & \textbf{2.18E-02} & - & 1.01E+01 & 4.05E+00 & = \\
 13 & \textbf{9.34E-29} & \textbf{2.14E-28} & \textbf{1.17E-28} & \textbf{2.34E-28} & = & 4.78E-01 & 1.30E+00 & + & \textbf{4.67E-29} & \textbf{1.58E-28} & = \\
 14 & 1.41E+00 & 0.00E+00 & 1.42E+00 & 2.36E-02 & = & \textbf{1.41E+00} & \textbf{0.00E+00} & - & 1.47E+00 & 4.05E-02 & + \\
 15 & \textbf{-3.93E+00} & \textbf{4.44E-16} & -3.93E+00 & 1.18E-05 & = & 2.36E+00 & 1.41E-06 & + & -3.42E+00 & 1.15E+00 & = \\
 16 & 1.99E+01 & 4.09E+00 & 2.10E+01 & 3.69E+00 & = & \textbf{0.00E+00} & \textbf{0.00E+00} & - & 2.32E+01 & 3.47E+00 & + \\
 17 & 7.54E-01 & 1.46E-01 & 8.37E-01 & 9.42E-02 & = & \textbf{6.81E-01} & \textbf{4.33E-01} & = & 8.41E-01 & 8.11E-02 & + \\
 18 & \textbf{3.65E+01} & \textbf{0.00E+00} & 3.65E+01 & 1.96E-05 & = & 1.85E+02 & 2.89E+01 & + & \textbf{3.65E+01} & \textbf{0.00E+00} & = \\
 19 & \textbf{0.00E+00} & \textbf{0.00E+00} & \textbf{0.00E+00} & \textbf{0.00E+00} & = & \textbf{0.00E+00} & \textbf{0.00E+00} & = & \textbf{0.00E+00} & \textbf{0.00E+00} & = \\
 20 & \textbf{1.26E+00} & \textbf{2.11E-01} & 1.33E+00 & 2.46E-01 & = & 1.85E+00 & 2.82E-01 & + & 2.43E+00 & 6.99E-01 & + \\
 21 & 2.16E+01 & 8.90E+00 & 2.39E+01 & 7.92E+00 & = & \textbf{9.28E+00} & \textbf{8.34E+00} & - & 2.47E+01 & 1.54E+01 & = \\
 22 & \textbf{3.75E-26} & \textbf{1.49E-26} & \textbf{5.61E-26} & \textbf{3.84E-26} & = & 2.56E+01 & 4.84E+01 & + & \textbf{2.85E-26} & \textbf{6.52E-27} & - \\
 23 & \textbf{1.41E+00} & \textbf{0.00E+00} & \textbf{1.41E+00} & \textbf{0.00E+00} & = & 1.45E+00 & 4.34E-02 & = & \textbf{1.41E+00} & \textbf{0.00E+00} & = \\
 24 & \textbf{-3.93E+00} & \textbf{4.44E-16} & \textbf{-3.93E+00} & \textbf{4.44E-16} & = & 2.36E+00 & 9.80E-08 & + & \textbf{-3.93E+00} & \textbf{4.44E-16} & = \\
 25 & 2.25E+01 & 3.00E+00 & 2.36E+01 & 3.69E+00 & = & \textbf{2.51E-01} & \textbf{1.23E+00} & - & 2.46E+01 & 3.10E+00 & = \\
 26 & 7.53E-01 & 1.50E-01 & 8.11E-01 & 7.52E-02 & = & \textbf{6.89E-01} & \textbf{2.63E-01} & = & 8.56E-01 & 1.18E-01 & + \\
 27 & 3.65E+01 & 8.06E-05 & 3.65E+01 & 8.00E-05 & = & 2.89E+02 & 3.21E+01 & + & \textbf{3.65E+01} & \textbf{2.71E-05} & - \\
 28 & \textbf{-5.12E+00} & \textbf{2.97E+00} & -4.75E+00 & 3.05E+00 & = & 7.72E+01 & 1.60E+01 & + & 0.00E+00 & 0.00E+00 & + \\
\hline
 W/T/L & \multicolumn{2}{c|}{$-/-/-$} & \multicolumn{3}{c|}{1/27/0} & \multicolumn{3}{c|}{14/7/7} & \multicolumn{3}{c|}{10/16/2} \\
\hline
\end{tabular}
\end{table}

\begin{table}[H]
\centering
\caption{Final constraint-violation comparison on the 28 CEC2025 CSOP functions. For each algorithm, the mean and SD over 25 runs are reported; $W$ gives the Wilcoxon outcome of RDEx-CSOP against the competitor.}
\label{tab:csop_cv_results}
\scriptsize
\renewcommand{\arraystretch}{0.9}
\setlength{\tabcolsep}{3pt}
\begin{tabular}{|c|cc|ccc|ccc|ccc|}
\hline
 \multirow{2}{*}{Problem} & \multicolumn{2}{c|}{RDEx-CSOP} & \multicolumn{3}{c|}{RDEx} & \multicolumn{3}{c|}{UDEIII} & \multicolumn{3}{c|}{CL-SRDE} \\ \cline{2-12}
   & Mean & SD & Mean & SD & W & Mean & SD & W & Mean & SD & W \\ \hline
 1 & \textbf{-2.70E+05} & \textbf{0.00E+00} & \textbf{-2.70E+05} & \textbf{0.00E+00} & = & 0.00E+00 & 0.00E+00 & + & \textbf{-2.70E+05} & \textbf{0.00E+00} & = \\
 2 & \textbf{-2.70E+05} & \textbf{0.00E+00} & \textbf{-2.70E+05} & \textbf{0.00E+00} & = & 0.00E+00 & 0.00E+00 & + & \textbf{-2.70E+05} & \textbf{0.00E+00} & = \\
 3 & \textbf{-1.99E+05} & \textbf{1.85E+04} & -1.85E+05 & 1.90E+04 & + & 0.00E+00 & 0.00E+00 & + & -1.52E+05 & 2.11E+04 & + \\
 4 & -4.45E+00 & 3.34E+00 & -5.01E+00 & 3.12E+00 & = & 0.00E+00 & 0.00E+00 & + & \textbf{-2.01E+01} & \textbf{3.74E+00} & - \\
 5 & \textbf{-2.14E+03} & \textbf{0.00E+00} & \textbf{-2.14E+03} & \textbf{0.00E+00} & = & 0.00E+00 & 0.00E+00 & + & \textbf{-2.14E+03} & \textbf{0.00E+00} & = \\
 6 & \textbf{0.00E+00} & \textbf{0.00E+00} & \textbf{0.00E+00} & \textbf{0.00E+00} & = & \textbf{0.00E+00} & \textbf{0.00E+00} & = & \textbf{0.00E+00} & \textbf{0.00E+00} & = \\
 7 & \textbf{0.00E+00} & \textbf{0.00E+00} & \textbf{0.00E+00} & \textbf{0.00E+00} & = & \textbf{0.00E+00} & \textbf{0.00E+00} & = & \textbf{0.00E+00} & \textbf{0.00E+00} & = \\
 8 & \textbf{0.00E+00} & \textbf{0.00E+00} & \textbf{0.00E+00} & \textbf{0.00E+00} & = & \textbf{0.00E+00} & \textbf{0.00E+00} & = & \textbf{0.00E+00} & \textbf{0.00E+00} & = \\
 9 & \textbf{-6.05E-37} & \textbf{2.80E-37} & \textbf{-4.91E-37} & \textbf{3.04E-37} & = & \textbf{0.00E+00} & \textbf{0.00E+00} & + & \textbf{-4.96E-37} & \textbf{3.07E-37} & = \\
 10 & \textbf{0.00E+00} & \textbf{0.00E+00} & \textbf{0.00E+00} & \textbf{0.00E+00} & = & \textbf{0.00E+00} & \textbf{0.00E+00} & = & \textbf{0.00E+00} & \textbf{0.00E+00} & = \\
 11 & 6.00E-10 & 1.51E-09 & 9.74E-11 & 4.67E-10 & - & 4.51E+01 & 1.22E+02 & + & \textbf{2.40E-15} & \textbf{1.16E-14} & - \\
 12 & -7.64E-01 & 1.49E-01 & \textbf{-7.94E-01} & \textbf{4.80E-05} & = & 0.00E+00 & 0.00E+00 & + & -7.67E-01 & 2.70E-01 & = \\
 13 & \textbf{-1.25E+02} & \textbf{0.00E+00} & \textbf{-1.25E+02} & \textbf{0.00E+00} & = & 0.00E+00 & 0.00E+00 & + & \textbf{-1.25E+02} & \textbf{0.00E+00} & = \\
 14 & \textbf{-1.00E+00} & \textbf{1.11E-16} & -9.20E-01 & 2.71E-01 & = & 0.00E+00 & 0.00E+00 & + & -3.20E-01 & 4.66E-01 & + \\
 15 & -1.37E+02 & 1.89E+02 & -1.73E+02 & 1.92E+02 & = & 0.00E+00 & 0.00E+00 & + & \textbf{-4.83E+02} & \textbf{2.65E+02} & - \\
 16 & \textbf{-2.97E+03} & \textbf{1.02E+01} & -2.97E+03 & 9.87E+00 & = & 0.00E+00 & 0.00E+00 & + & -2.97E+03 & 1.12E+01 & = \\
 17 & 3.10E+01 & 0.00E+00 & 3.10E+01 & 0.00E+00 & = & \textbf{2.94E+01} & \textbf{7.89E-01} & - & 3.10E+01 & 0.00E+00 & = \\
 18 & \textbf{-3.00E+03} & \textbf{4.55E-13} & \textbf{-3.00E+03} & \textbf{4.55E-13} & = & 7.75E+03 & 2.49E+03 & + & \textbf{-3.00E+03} & \textbf{4.55E-13} & = \\
 19 & \textbf{4.27E+04} & \textbf{0.00E+00} & \textbf{4.27E+04} & \textbf{0.00E+00} & = & 4.27E+04 & 7.28E-12 & + & \textbf{4.27E+04} & \textbf{0.00E+00} & = \\
 20 & -3.66E-01 & 1.81E-01 & -2.89E-01 & 1.81E-01 & = & 0.00E+00 & 0.00E+00 & + & \textbf{-3.87E-01} & \textbf{1.73E-01} & = \\
 21 & -1.32E+00 & 3.97E-01 & \textbf{-1.42E+00} & \textbf{3.53E-01} & = & 0.00E+00 & 0.00E+00 & + & -1.33E+00 & 6.25E-01 & = \\
 22 & \textbf{-1.25E+02} & \textbf{0.00E+00} & \textbf{-1.25E+02} & \textbf{0.00E+00} & = & 0.00E+00 & 0.00E+00 & + & \textbf{-1.25E+02} & \textbf{0.00E+00} & = \\
 23 & \textbf{-1.00E+00} & \textbf{1.11E-16} & \textbf{-1.00E+00} & \textbf{1.11E-16} & = & 0.00E+00 & 0.00E+00 & + & \textbf{-1.00E+00} & \textbf{1.11E-16} & = \\
 24 & -7.38E+01 & 1.00E+02 & -8.14E+01 & 1.36E+02 & = & 0.00E+00 & 0.00E+00 & + & \textbf{-1.98E+02} & \textbf{2.32E+02} & = \\
 25 & \textbf{-2.97E+03} & \textbf{8.21E+00} & -2.97E+03 & 1.04E+01 & = & 0.00E+00 & 0.00E+00 & + & -2.97E+03 & 8.66E+00 & = \\
 26 & 3.10E+01 & 0.00E+00 & 3.10E+01 & 0.00E+00 & = & \textbf{3.06E+01} & \textbf{8.09E-01} & = & 3.10E+01 & 0.00E+00 & = \\
 27 & \textbf{-3.00E+03} & \textbf{4.55E-13} & \textbf{-3.00E+03} & \textbf{4.55E-13} & = & 1.43E+04 & 5.11E+03 & + & \textbf{-3.00E+03} & \textbf{4.55E-13} & = \\
 28 & 4.28E+04 & 1.14E+01 & 4.28E+04 & 1.10E+01 & = & 4.29E+04 & 1.47E+01 & + & \textbf{4.27E+04} & \textbf{0.00E+00} & - \\
\hline
 W/T/L & \multicolumn{2}{c|}{$-/-/-$} & \multicolumn{3}{c|}{1/26/1} & \multicolumn{3}{c|}{22/5/1} & \multicolumn{3}{c|}{2/22/4} \\
\hline
\end{tabular}
\end{table}

\begin{table}[H]
\centering
\caption{Anytime convergence comparison using AUC over 2000 checkpoints on the 28 CEC2025 CSOP functions. For each run, AUC is computed as the mean of $\log_{10}(1+\max(f_t-\mathrm{TGT},0))$ across checkpoints (smaller is better).}
\label{tab:csop_auc_results}
\scriptsize
\renewcommand{\arraystretch}{0.9}
\setlength{\tabcolsep}{3pt}
\begin{tabular}{|c|cc|ccc|ccc|ccc|}
\hline
 \multirow{2}{*}{Problem} & \multicolumn{2}{c|}{RDEx-CSOP} & \multicolumn{3}{c|}{RDEx} & \multicolumn{3}{c|}{UDEIII} & \multicolumn{3}{c|}{CL-SRDE} \\ \cline{2-12}
   & Mean & SD & Mean & SD & W & Mean & SD & W & Mean & SD & W \\ \hline
 1 & 0.17 & 0.01 & 0.17 & 0.00 & = & \textbf{0.14} & \textbf{0.01} & - & 0.32 & 0.01 & + \\
 2 & 0.20 & 0.01 & 0.20 & 0.01 & = & \textbf{0.14} & \textbf{0.01} & - & 0.38 & 0.01 & + \\
 3 & 1.87 & 0.65 & 2.18 & 0.50 & = & \textbf{0.50} & \textbf{0.18} & - & 3.15 & 0.11 & + \\
 4 & \textbf{0.78} & \textbf{0.24} & 0.84 & 0.32 & = & 0.81 & 0.04 & = & 1.80 & 0.08 & + \\
 5 & 0.36 & 0.01 & 0.36 & 0.01 & = & \textbf{0.30} & \textbf{0.01} & - & 0.60 & 0.02 & + \\
 6 & \textbf{0.27} & \textbf{0.01} & 0.39 & 0.33 & = & 0.52 & 0.02 & + & 0.62 & 0.30 & + \\
 7 & \textbf{0.74} & \textbf{0.66} & 1.34 & 0.86 & + & 2.30 & 0.47 & + & 2.00 & 0.61 & + \\
 8 & 0.10 & 0.00 & \textbf{0.10} & \textbf{0.00} & = & 0.33 & 0.06 & + & 0.20 & 0.01 & + \\
 9 & 0.02 & 0.00 & \textbf{0.02} & \textbf{0.00} & = & 0.06 & 0.04 & + & 0.06 & 0.01 & + \\
 10 & \textbf{0.10} & \textbf{0.00} & 0.10 & 0.00 & = & 0.39 & 0.05 & + & 0.19 & 0.00 & + \\
 11 & \textbf{0.02} & \textbf{0.04} & 0.04 & 0.05 & = & 0.04 & 0.06 & = & 0.10 & 0.03 & + \\
 12 & 0.30 & 0.01 & 0.30 & 0.01 & = & \textbf{0.26} & \textbf{0.03} & - & 0.56 & 0.20 & + \\
 13 & 0.70 & 0.02 & 0.70 & 0.02 & = & \textbf{0.59} & \textbf{0.21} & - & 1.16 & 0.04 & + \\
 14 & \textbf{0.10} & \textbf{0.01} & 0.10 & 0.01 & = & 0.11 & 0.02 & + & 0.30 & 0.10 & + \\
 15 & \textbf{0.06} & \textbf{0.01} & 0.06 & 0.01 & = & 0.90 & 0.02 & + & 0.30 & 0.17 & + \\
 16 & 0.54 & 0.33 & 0.63 & 0.32 & = & \textbf{0.12} & \textbf{0.07} & - & 1.06 & 0.18 & + \\
 17 & \textbf{0.02} & \textbf{0.01} & 0.03 & 0.02 & = & 0.05 & 0.04 & = & 0.05 & 0.01 & + \\
 18 & 0.27 & 0.02 & \textbf{0.27} & \textbf{0.01} & = & 2.18 & 0.06 & + & 0.43 & 0.01 & + \\
 19 & 0.23 & 0.01 & 0.24 & 0.02 & = & \textbf{0.09} & \textbf{0.03} & - & 0.37 & 0.01 & + \\
 20 & \textbf{0.33} & \textbf{0.04} & 0.34 & 0.04 & = & 0.39 & 0.05 & + & 0.61 & 0.06 & + \\
 21 & \textbf{0.32} & \textbf{0.03} & 0.32 & 0.02 & = & 0.42 & 0.21 & + & 0.70 & 0.40 & + \\
 22 & \textbf{0.82} & \textbf{0.04} & 0.83 & 0.04 & = & 1.51 & 0.46 & + & 1.35 & 0.03 & + \\
 23 & \textbf{0.11} & \textbf{0.00} & 0.12 & 0.00 & = & 0.19 & 0.02 & + & 0.19 & 0.00 & + \\
 24 & \textbf{0.09} & \textbf{0.01} & 0.09 & 0.02 & = & 0.91 & 0.02 & + & 0.26 & 0.02 & + \\
 25 & 0.55 & 0.23 & 0.63 & 0.26 & = & \textbf{0.43} & \textbf{0.40} & - & 1.01 & 0.15 & + \\
 26 & 0.03 & 0.01 & 0.04 & 0.01 & = & \textbf{0.03} & \textbf{0.02} & = & 0.08 & 0.02 & + \\
 27 & 0.31 & 0.03 & \textbf{0.31} & \textbf{0.02} & = & 2.43 & 0.04 & + & 0.48 & 0.01 & + \\
 28 & \textbf{0.20} & \textbf{0.02} & 0.20 & 0.02 & = & 1.98 & 0.06 & + & 0.38 & 0.01 & + \\
\hline
 W/T/L & \multicolumn{2}{c|}{$-/-/-$} & \multicolumn{3}{c|}{1/27/0} & \multicolumn{3}{c|}{15/4/9} & \multicolumn{3}{c|}{28/0/0} \\
\hline
\end{tabular}
\end{table}

\end{CJK}
\end{document}